\documentclass[letterpaper]{article} 
\usepackage[preprint]{aaai2027}  
\usepackage[hyphens]{url}  
\usepackage{graphicx} 
\usepackage{natbib}  
\usepackage{caption} 
\usepackage{amsmath}
\usepackage{amssymb}
\usepackage{booktabs}
\usepackage{array}
\usepackage{tabularx}
\usepackage{multirow}
\usepackage{tikz}
\usetikzlibrary{arrows.meta,positioning,fit,calc}

\newcolumntype{Y}{>{\raggedright\arraybackslash}X}
\newcolumntype{Z}{>{\centering\arraybackslash}X}
\newcommand{\ours}{Hi-OPD}
\newcommand{\bench}{RS153-HierOPD}

\title{Hi-OPD: Hierarchy-Aware Open-Prompt Detection for Remote Sensing Images}
\author{
	Jinlong Hu\textsuperscript{\rm 1},
	Yi Zhang\textsuperscript{\rm 2},
	Zhiqi Xia\textsuperscript{\rm 1},
	Yikang Zhou\textsuperscript{\rm 1},
	Shunping Ji\textsuperscript{\rm 1}
}
\affiliations{
	\textsuperscript{\rm 1}Wuhan University\\
	\textsuperscript{\rm 2}Institute of Seismology, China Earthquake Administration
}

\begin{document}
	
	\maketitle
	
	\begin{abstract}
		\ours{} addresses a failure mode left uncontrolled by flat open-prompt training: descendant retrieval need not persist under ancestor queries when multi-source remote sensing annotations exhibit inconsistent granularity and missing labels. A detector may localize \textit{car} and \textit{van} under atomic prompts yet miss the same instances under \textit{vehicle}; flat AP does not expose this cross-level inconsistency.

		We propose \ours, a hierarchy-aware open-prompt detector, and construct \bench{} from 175,644 retained training image/tile records and 3.48M boxes mapped to 153 atomic categories with sparse hierarchy and alias relations. \ours{} learns ancestor retrieval through hierarchy-safe negative sampling, path multi-positive supervision, and one-way upward consistency, while per-source risk exclusion handles potentially missing labels. ConvVPE converts K-shot support boxes into text-compatible embeddings using detector-native features and the shared contrastive head.

		On Track A, \ours{} obtains 79.7/72.3 AP50 on DIOR/DOTA-v2.0, above the literature-reported OpenRSD results of 76.7/71.8. Under controlled training on the original converted annotations, the full hierarchy recipe raises DOTA-v2.0 parent AP50 from 7.2 to 71.5 and FAIR1M grandparent AP50 from 31.6 to 71.4, while DOTA-v2.0 atomic AP50 changes from 71.4 to 72.3. The text path reaches 99.7\% CAR50 (0.3\% violation) across the three common sources and 99.9\%/0.1\% on FAIR1M grandparent relations. On held-out VEDAI, text AP50 is 75.9, 6.2 points above OpenRSD. Joint AP and CAR show that explicit hierarchy training repairs this failure mode while retaining atomic detection and prompt transfer.
	\end{abstract}
	
	\section{Introduction}
	
	Object detection in remote sensing images underpins applications such as urban monitoring, traffic analysis, disaster response, and land-use management. Most remote sensing benchmarks assume a fixed category set shared by training and testing, an assumption strained by new areas, sensors, tasks, and annotation policies.
	
	Open-prompt detection relaxes this assumption by accepting category prompts at inference time. Recent remote sensing systems such as OpenRSD~\cite{ref1} and LAE-DINO~\cite{ref2} extend this paradigm with visual prompting and domain-specific training, demonstrating strong atomic-category detection and cross-dataset transfer.

	\begin{figure}[!t]
		\centering
		\includegraphics[width=\linewidth,height=0.27\textheight,keepaspectratio]{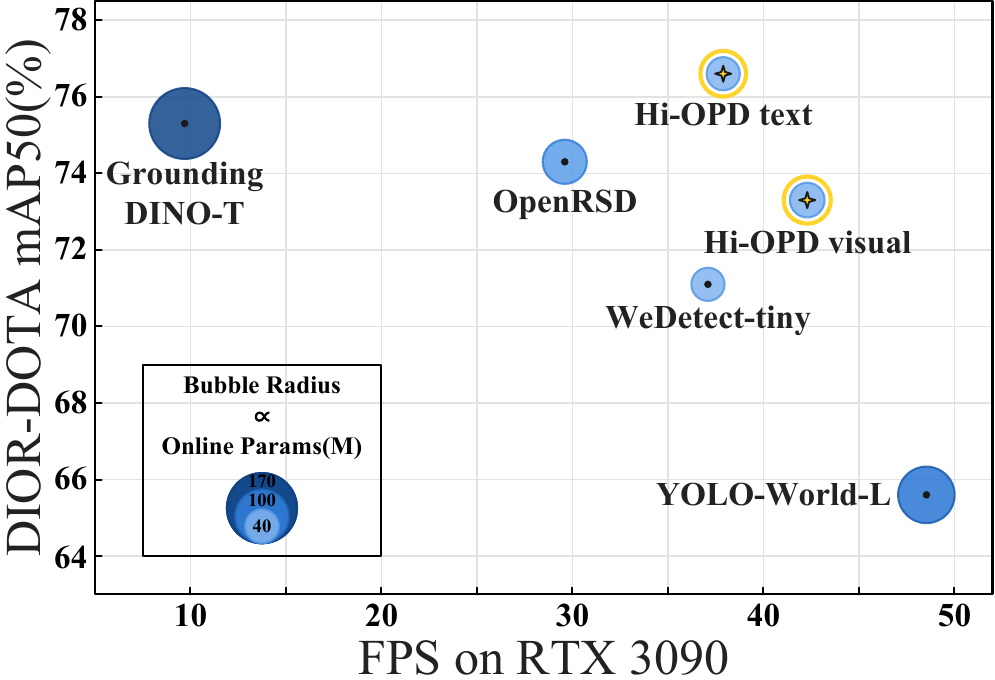}
		\caption{Accuracy--efficiency--size trade-off. The axes report mean DIOR/DOTA-v2.0 AP50 and FP32 FPS on an RTX 3090; bubble radius is proportional to online/evaluation parameters. Stars denote \ours{} variants. LAE-DINO is omitted because DOTA AP50 is not reported.}
		\label{fig:speed_positioning}
	\end{figure}
	
	Despite this progress, existing remote sensing open-prompt systems largely organize prompts as \textit{flat category labels}: query a class name, detect that class, and report average precision over a flat category set. Strong atomic detection therefore does not guarantee that the same objects remain retrievable when a user issues a coarser ancestor prompt.

	Semantic hierarchies have been explored in general-domain open-vocabulary detection. SHiNe~\cite{Liu2024_SHiNe} constructs training-free hierarchy-aware classifiers from super- and sub-category descriptions; LHST~\cite{Huang2024_LHST} introduces language hierarchies into weakly supervised self-training and prompt generation; and HCC~\cite{Lee2026_HCC} uses consistency across class, super-category, and sub-category predictions to calibrate pseudo-label confidence. These studies establish that vocabulary granularity matters, but focus on classifier construction, image-level weak supervision, or pseudo-label calibration. They do not study direct ancestor-query retrieval under heterogeneous remote sensing annotations, where category granularity and missing-box patterns vary by source.

	Remote sensing couples two challenges that flat prompting leaves unresolved: hierarchy-dependent queries and source-dependent partial labels.
	
	\textbf{Missing hierarchical recall.}
	Operational search is often coarse-to-fine, yet flat AP does not test whether objects detected under fine-grained labels remain retrievable from a parent prompt. Track B directly measures this capability.
	
	\textbf{Multi-source granularity conflicts.}
	Sources differ in annotation granularity, and single- or few-class datasets leave co-visible objects outside their target taxonomies unlabeled. Treating those categories as negatives suppresses true objects, whereas merging parent and child labels erases fine-grained distinctions.
	
	We therefore ask how the same object instances can be retrieved consistently across semantic granularities without introducing false negatives from heterogeneous annotations. \ours{} addresses this question with a unified 153-class atomic vocabulary, a sparse explicit hierarchy over semantically meaningful subsets, hierarchy-safe text training, and detector-native visual prompting. The hierarchy is intentionally not exhaustive: only atomic categories with operationally meaningful ancestor queries participate in the corresponding hierarchy evaluations.
	
	Our contributions are fourfold:
	\begin{itemize}
		\item We introduce \bench, which unifies multi-source remote sensing data into 153 atomic categories and records explicit parent--child and alias relations over hierarchy-enabled subsets. It supports atomic detection, direct hierarchy-query evaluation with both absolute AP and conditional ancestor recall, and cross-dataset prompt generalization.
		\item We formulate direct ancestor-query detection under heterogeneous annotation granularity. \ours{} prevents semantically valid prompts from becoming false negatives through hierarchy-safe negative sampling, path multi-positive supervision, and one-way upward consistency.
		\item To enable lightweight, detector-native visual prompting, we propose ConvVPE, which encodes K-shot support regions directly from the detector's own multi-scale features into visual prompt embeddings compatible with the text-prompt space, thereby reusing the same detection head without introducing an external visual encoder.
		\item We provide controlled evaluation across atomic detection, direct parent and grandparent queries, and held-out cross-dataset transfer. Within the same detector architecture and training protocol, hierarchy-aware training repairs the ancestor-query failure mode of flat training while preserving atomic detection.
	\end{itemize}
	
	\section{Related Work}
	
	\textbf{Open-prompt detection.}
	Open-prompt detection casts category recognition as matching between region features and category embeddings. CLIP~\cite{ref4} provides transferable vision-language representations; GLIP~\cite{ref5} and Grounding DINO~\cite{ref6} introduce phrase grounding into detection pretraining; other representative systems~\cite{ref7,ref8,ref9,ref10,ref11,ref12,ref13} span vision-transformer open-vocabulary detection, region--language distillation and pretraining, text-conditioned detection, image-level supervision, and efficient deployment. WeDetect~\cite{ref40} reformulates open-category detection as retrieval and provides a fast YOLO-World-style detector; we use WeDetect-tiny as the base detector and build both text and visual prompt paths on top of it.
	
	\textbf{Remote sensing detection and open-prompt detection.}
	Remote sensing detection differs from natural image detection due to overhead viewpoints, dense small objects, large scale variation, and specialized category names. DOTA~\cite{ref15}, DIOR~\cite{ref16}, FAIR1M~\cite{ref17}, HRSC2016~\cite{ref21}, ShipRSImageNet~\cite{ref22}, and SODA-A~\cite{ref20} have advanced horizontal, oriented, and fine-grained aerial object detection. OpenRSD~\cite{ref1} systematically introduces text- and image-prompt detection in remote sensing, while LAE-DINO~\cite{ref2} builds a large remote sensing corpus and dynamic category set to reduce domain mismatch. Their reported protocols primarily evaluate flat category vocabularies. \ours{} retains atomic and cross-dataset evaluation while adding direct ancestor queries and explicit treatment of multi-source granularity conflicts.

	\textbf{Hierarchy-aware open-vocabulary detection.}
	Earlier work uses WordNet~\cite{ref32}, WordTree detectors~\cite{ref33}, and hierarchy-aware objectives~\cite{ref39,ref49} to organize labels or trade accuracy against semantic specificity. More recently, SHiNe~\cite{Liu2024_SHiNe} exposes the instability of open-vocabulary detectors across vocabulary granularities and builds a training-free nexus classifier by integrating super- and sub-category descriptions. LHST~\cite{Huang2024_LHST} expands image-level labels through a language hierarchy and co-regularizes weakly supervised self-training, whereas HCC~\cite{Lee2026_HCC} calibrates pseudo-label confidence from class/super-category/sub-category consistency. These general-domain methods use hierarchy for classifier construction, weak image-level supervision, or pseudo-label selection. In contrast, \ours{} directly supervises ancestor prompts for descendant boxes and prevents ancestor, descendant, alias, and source-risk prompts from becoming false negatives under heterogeneous remote sensing annotations.
	
	\textbf{Multi-dataset taxonomies and incomplete labels.}
	Unified-label detectors~\cite{ref45}, Simple Multi-Dataset Detection~\cite{ref46}, Detection Hub~\cite{ref47}, and ScaleDet~\cite{ref48} address heterogeneous label spaces, semantic alignment, and domain variation. LVIS~\cite{ref34}, Detic~\cite{ref10}, and long-tail losses~\cite{ref37,ref38} address vocabulary scale, weak supervision, or class imbalance. These approaches do not directly resolve the coupled case in which one source uses a coarse parent label, another uses its fine-grained descendants, and non-target objects remain unannotated. \ours{} uses explicit hierarchy, alias, and source-risk relations to determine which prompts are valid positives, unsafe negatives, or masked categories.
	
	\textbf{Visual prompting.}
	Visual prompts specify categories through examples rather than names. T-Rex2~\cite{ref26} supports box- or point-conditioned prompts, while OpenRSD~\cite{ref1} extracts remote sensing image prompts with an external DINOv2 encoder~\cite{ref28}. ConvVPE instead pools support regions from the detector's own multi-scale features and maps them into the shared contrastive prompt space, providing a lightweight visual extension without a separate visual encoder.

	\section{\bench{} Benchmark}
	
	\bench{} maps twelve heterogeneous sources into 153 atomic categories. Its submitted manifests contain 175,644 training image/tile records with 3,477,064 boxes and 56,695 validation records with 1,297,548 boxes. The hierarchy is sparse rather than exhaustive: 108 atomic categories participate in 108 direct and 70 grandparent relations, using 11 parent and two grandparent nodes. Eleven atomic classes also serve as ancestors, and three alias groups provide 11 protected links.

	Construction retains visually decidable atomic categories, uses strict \textit{is-a} relations for hierarchy edges, and reserves alias links for naming equivalence or protocol-level ambiguity. Part-of relations, co-occurrence, and appearance similarity alone are excluded. Table~\ref{tab:rs153_stats} makes the evaluation-bearing structure explicit. The code-and-data appendix includes the complete taxonomy, all hierarchy and alias edges, all 246 source-label mappings, effective per-source Risk sets, split provenance, relation-level evaluation metadata, audit boundaries, and validation scripts.

	\begin{center}
		\centering
		\scriptsize
		\setlength{\tabcolsep}{2pt}
		\begin{tabular}{@{}lr@{}}
			\toprule
			Benchmark item & Count \\
			\midrule
			Atomic / hierarchy-enabled atomic classes & 153 / 108 \\
			Parent / grandparent nodes & 11 / 2 \\
			Direct / grandparent relations & 108 / 70 \\
			Alias groups / protected links & 3 / 11 \\
			Atomic classes also serving as ancestors & 11 \\
			Train records / boxes & 175,644 / 3,477,064 \\
			Validation records / boxes & 56,695 / 1,297,548 \\
			Risk-covered sources (global / source-specific) & 12 / 3 \\
			Track-B relations (common / standalone GP) & 36 / 9 \\
			Track-B GT (common / standalone GP) & 782,132 / 290,906 \\
			\bottomrule
		\end{tabular}
		\captionof{table}{Review-time \bench{} statistics. Track-B counts use source-specific relation entries; the FAIR1M depth-2 diagnostic is separate.}
		\label{tab:rs153_stats}
	\end{center}
	
	\bench{} defines three tasks. Track A measures atomic detection over the 153-class vocabulary. Track B rolls ground truth to the requested level and provides only the corresponding coarse prompts; its common direct/mixed protocol contains 36 source-specific relations and 782,132 GT instances, while nine FAIR1M depth-2 relations (290,906 GT) form a separate grandparent diagnostic. Track C evaluates zero-shot text and K-shot visual prompt generalization on datasets excluded from training. A visual summary of the three evaluation tasks is provided in the technical appendix.
		
	\section{Method}
	
	\subsection{Overview}
	
	\ours{} addresses two coupled problems in multi-source remote sensing open-prompt detection: category hierarchy and source-dependent partial labels. Given an image $I$ and a prompt category set $C$, the detector extracts region features $r_i$ and matches each region to a prompt embedding $z_c$:
	\begin{equation}
		s_{ic}=\gamma\, r_i^\top z_c + b_c ,
	\end{equation}
	where $\gamma$ is a learnable logit scale and $b_c$ is a category bias. Unlike flat prompt training, \ours{} decides which classes can safely serve as negatives using parent, descendant, alias, and source-risk relations.
	
	\begin{figure*}[t]
		\centering
		\includegraphics[width=0.98\textwidth]{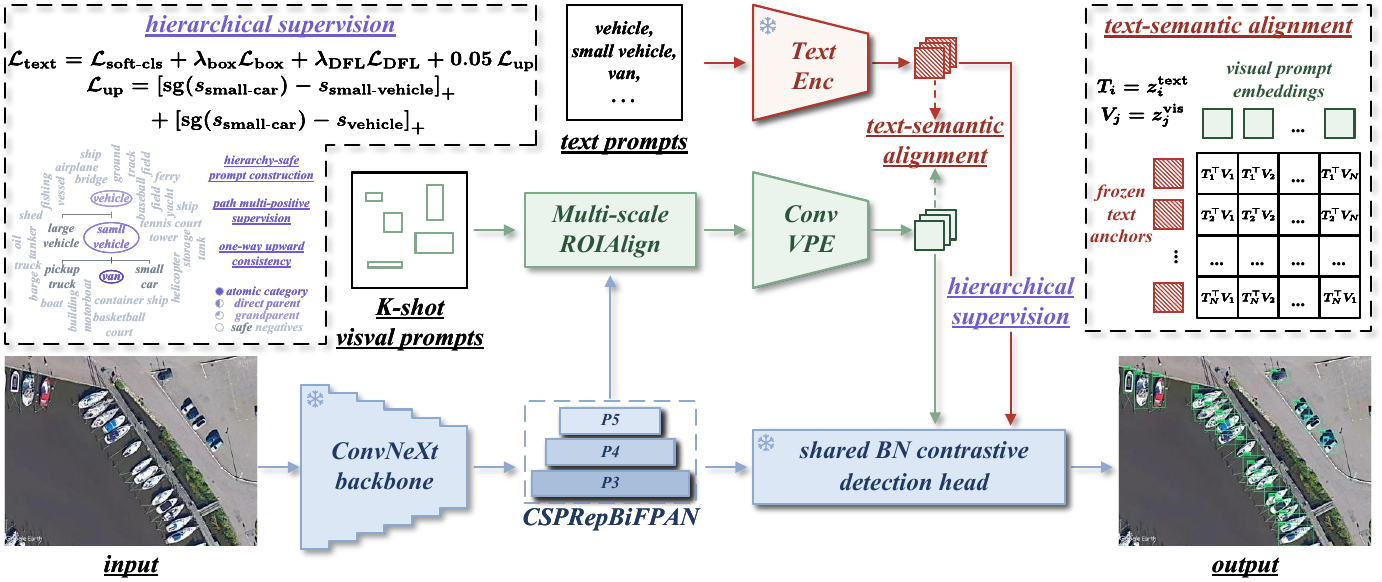}
		\caption{Overview of \ours. \bench{} provides multi-source data and hierarchy/alias metadata. The text detector is trained through staged hierarchy-aware continuation with path multi-positive supervision and one-way upward consistency. ConvVPE is trained separately as a frozen-body visual prompt branch. Text and visual prompts share the same BN contrastive detection head.}
		\label{fig:overview}
	\end{figure*}
	
	Figure~\ref{fig:overview} illustrates the framework. \ours{} has two prompt paths. The text path performs prompt-based inference from category names without requiring support examples, whereas the visual path uses ConvVPE to construct category prompts from K-shot support boxes. Both paths produce 768-dimensional prompt embeddings compatible with the same batch-normalized (BN) contrastive detection head:
	\[
	z_c =
	\begin{cases}
		\mathrm{Normalize}(T(c)), & \text{text},\\
		\mathrm{Normalize}(\mathrm{ConvVPE}(S_c;\phi)), & \text{visual}.
	\end{cases}
	\]
	Here $T(\cdot)$ is the cached text encoder, $S_c$ is the support set of class $c$, and $\phi$ denotes ConvVPE parameters.
	
	\subsection{Hierarchy-Safe Prompt Construction}
	
	Flat negative sampling can generate false-negative supervision. For a region of class $y$, its parent prompt should also be able to recall this region. If the parent, a descendant, an alias, or a source-risk class is sampled as a negative, training explicitly suppresses a semantically valid prompt. Let $\mathcal{H}(Y)=\mathrm{Anc}(Y)\cup\mathrm{Desc}(Y)\cup\mathrm{Alias}(Y)$ denote the hierarchy/alias neighborhood of observed classes. We define the potential false-negative pool as
	\begin{equation}
		F(Y,s)=\big(\mathcal{H}(Y)\cup \mathrm{Risk}(s)\big)\setminus Y ,
	\end{equation}
	where $Y$ is the set of observed positive classes in a source-$s$ image, and $\mathrm{Anc}$, $\mathrm{Desc}$, and $\mathrm{Alias}$ denote ancestor, descendant, and alias sets. $\mathrm{Risk}(s)$ is an unsafe-negative set rather than an additional positive-label set: it contains categories that are plausible in source $s$ but are not exhaustively annotated under that source's labeling policy. The executable policy contains one global protection rule applied to all twelve sources and additional source-specific rules for three sources. It has two uses. First, during headline \ours{} training, its members are excluded from the sampled negative prompt pool so that potentially unlabeled instances are not explicitly suppressed. Second, in a separate auxiliary experiment, the responses preserved for $\mathrm{Risk}(s)$ categories are used to mine candidates for supplemental annotation. The supplemental annotations are not used to train the headline models; this second use evaluates data-curation potential rather than a separate final-AP module. The technical appendix reports the candidate-mining analysis, and the code-and-data appendix includes the complete effective set for every source.
	
	The hierarchy-safe negative set is
	\begin{equation}
		N_{\mathrm{safe}} = C \setminus \big(Y \cup \mathcal{H}(Y)\cup \mathrm{Risk}(s)\big).
	\end{equation}
	The training prompt pool contains observed positive classes, ancestor prompts along the semantic path, and sampled safe negatives:
	\begin{equation}
		P_{\mathrm{train}} = Y \cup \mathrm{Anc}_{\leq 2}(Y) \cup \mathrm{Sample}(N_{\mathrm{safe}}).
	\end{equation}
	For a region whose atomic category is $y_i$, we supervise the whole valid hierarchy path rather than only the leaf. Let $\Pi_i=\{y_i\}\cup\mathrm{Anc}_{\leq 2}(y_i)$ be the path from the atomic category to its direct parent and grandparent. The path multi-positive target is
	\begin{equation}
		t_{ic}=
		\begin{cases}
			1, & c=y_i,\\
			\beta_d, & c\in \mathrm{Anc}_d(y_i),~d\leq 2,\\
			0, & c\in N_{\mathrm{safe}},
		\end{cases}
	\end{equation}
	where $\beta_d$ is an ancestor-positive weight. We set $\beta_1=\beta_2=0.8$, $m=0$, and $\lambda_{\mathrm{up}}=0.05$ in the final recipe. Thus, both direct parents and selected grandparents receive the same path-positive weight. This differs from post-hoc label roll-up: the detector learns to score each hierarchy prompt directly, matching the hierarchy-query protocol used at evaluation time.
	
	Path supervision alone does not explicitly order logits along the hierarchy. We therefore add a one-way upward consistency term:
	\begin{equation}
		\mathcal{L}_{\mathrm{up}}
		= \frac{1}{|\mathcal{P}|}
		\sum_{(i,a)\in\mathcal{P}}
		\left[
		\mathrm{sg}(s_{i y_i}) - s_{ia} + m
		\right]_+ ,
	\end{equation}
	where $a\in\mathrm{Anc}_{\leq 2}(y_i)$, $\mathrm{sg}(\cdot)$ stops gradients through the child logit (\texttt{detach\_child=True}), and $m$ is a margin. Because $m=0$, this is a zero-margin one-way ordering constraint: each ancestor logit is required to be no lower than the detached child logit. The loss can only pull ancestor logits upward when they fall below the atomic evidence; it does not push down fine-grained atomic logits. This one-way design is important for improving parent and grandparent recall while preserving Track-A atomic detection. Classes not covered by the target or safe negatives are masked from the classification loss. The final text-path detection objective is
	\begin{equation}
		\mathcal{L}_{\mathrm{det}}
		= \mathcal{L}_{\mathrm{cls}}(s_{ic},t_{ic})
		+ \lambda_{\mathrm{box}}\mathcal{L}_{\mathrm{box}}
		+ \lambda_{\mathrm{dfl}}\mathcal{L}_{\mathrm{dfl}}
		+ \lambda_{\mathrm{up}}\mathcal{L}_{\mathrm{up}} .
	\end{equation}

	\subsection{Detector-Native Visual Prompt Encoder}
	
	Existing image-prompt methods often rely on an external visual encoder to extract support features. ConvVPE instead runs on the detector's own multi-scale features. Given class $c$ and support set $S_c=\{(I_m,b_m)\}_{m=1}^{M}$, ConvVPE applies multi-scale ROI alignment on detector feature maps and augments the crop with box-coordinate encoding:
	\begin{equation}
		u_m=\mathrm{ROIAlign}\big(\{P_3,P_4,P_5\}, b_m\big)\oplus e(b_m).
	\end{equation}
	A lightweight convolutional encoder and projection layer map each support sample to a visual anchor:
	\begin{equation}
		\hat{z}_{m} = \mathrm{Proj}\big(\mathrm{ConvEncoder}(u_m)\big).
	\end{equation}
	For multiple support samples of the same class, a gating branch estimates normalized support weights:
	\begin{equation}
		\begin{aligned}
			a_m&=\frac{\exp(g(\hat{z}_m))}{\sum_{j=1}^{M}\exp(g(\hat{z}_j))},\\
			z_c^{\mathrm{vis}}&=\mathrm{Normalize}\Big(\sum_{m=1}^{M}a_m\hat{z}_m\Big).
		\end{aligned}
	\end{equation}
	ConvVPE only aggregates support samples within a class; it does not perform text-visual fusion. The resulting visual embedding has the same dimensionality as the text embedding and is directly consumed by the shared detection head.
	
	\subsection{Training and Inference}
	
	\textbf{Text detector training stages.}
	Stage 1 learns the multi-source text detector, and Stage 2 continues it with the hierarchy-aware objective above. The resulting checkpoint is used for text-prompt inference and initializes the frozen-body ConvVPE branch.
	
	\textbf{Frozen-body visual prompt learning.}
	ConvVPE is initialized from the Stage-2 hierarchy-aware text checkpoint and trained as an independent visual-prompt branch. During this step, the detector body---including the backbone, neck, non-BN detection-head parameters, and cached text embeddings---is frozen, and only ConvVPE parameters are updated. For K-shot support boxes, ConvVPE maps support samples to 768-dimensional prompt embeddings, which are normalized and fed into the same BN contrastive head as text embeddings. The training objective is
	\begin{equation}
		\mathcal{L}_{\mathrm{vis}}
		=\mathcal{L}_{\mathrm{cls}}^{\mathrm{vis}}+\lambda_{\mathrm{align}}\mathcal{L}_{\mathrm{align}},
	\end{equation}
	where $\mathcal{L}_{\mathrm{cls}}^{\mathrm{vis}}$ uses the same hierarchy-safe classification loss through the shared head, and $\mathcal{L}_{\mathrm{align}}$ aligns visual prompt embeddings with their corresponding text anchors. We set $\lambda_{\mathrm{align}}=1.0$. During training, the number of supports is sampled from $M\in\{1,2,3,5\}$ with support dropout; evaluation uses $M=5$.
	
	\begin{table*}[!t]
		\centering
		\footnotesize
		\resizebox{\textwidth}{!}{%
			\begin{tabular}{@{}llccccccc@{}}
				\toprule
				Method & Prompt & \multicolumn{3}{c}{Track A: atomic} & \multicolumn{2}{c}{Track B: hierarchy} & \multicolumn{2}{c}{Efficiency} \\
				\cmidrule(lr){3-5}\cmidrule(lr){6-7}\cmidrule(lr){8-9}
				& & DIOR AP50 & DOTA mAP & DOTA AP50 & Parent-child AP50 & Grandparent-child AP50 & FPS & Params (M) \\
				\midrule
				\multicolumn{9}{@{}l}{\textit{Panel A: RS153-HierOPD evaluations in this work}} \\
				WeDetect-tiny (zero-shot) & text & 6.2 & 2.3 & 4.9 & 9.3 / 11.5 & 13.0 & 37.1 & 38.1 \\
				WeDetect-tiny (per-source ft) & text & 76.5 & 42.6 & 65.6 & 50.5 / 50.2 & 30.5 & 37.1 & 38.1 \\
				\textbf{\ours{} text} & hier. text & 79.7 & \textbf{48.2} & \textbf{72.3} & \textbf{81.1 / 76.0} & \textbf{71.4} & 37.9 & \textbf{38.1} \\
					\ours{} visual & visual 5-shot & 75.2 & 46.1 & 69.5 & 61.9 / 59.7 & 57.2 & 36.4 & 41.5 \\
				\midrule
				\multicolumn{9}{@{}l}{\textit{Panel B: external methods (method-native training and prompts)}} \\
				YOLO-World-L~\cite{ref13}$^\star$ & text & 73.2 & -- & 58.0 & 35.5 / 36.5 & 22.5 & \textbf{48.5} & 110.3 \\
				Grounding DINO-T~\cite{ref6}$^\star$ & text & 78.7 & -- & 71.8 & 35.6 / 38.3 & 53.2 & 9.7 & 173.0 \\
				OpenRSD~\cite{ref1} & text & 76.7 & -- & 71.8 & 57.0 / 50.5 & 60.8 & 29.6 & 67.2 \\
				LAE-DINO~\cite{ref2} & text & \textbf{85.5}$^\ddagger$ & 46.8 & -- & 33.6 / 27.5 & 20.9 & 7.9 & 181.6 \\
				\bottomrule
		\end{tabular}}
		\caption{Main Track-A and Track-B results. Panel~A reports evaluations in this work. Panel~B combines literature-reported Track-A values with Track-B evaluations in this work; sources and protocols are detailed in the setup. $^\star$For Track B, YOLO-World-L and Grounding DINO-T are flat-fine-tuned separately on each source for 10 epochs. Parent-child AP50 is shown as common/all-source macro, and grandparent-child AP50 is the FAIR1M \textit{ship}/\textit{vehicle} diagnostic. $^\ddagger$LAE-DINO DIOR follows the broader seen-data protocol in its original paper.}
		\label{tab:main_eval}
	\end{table*}

	\section{Experiments}
	
	\subsection{Experimental Setup}
	
	\textbf{Data.}
	The \bench{} training set comprises twelve sources: DIOR~\cite{ref16}, DOTA-v2.0~\cite{ref15}, FAIR1M~\cite{ref17}, GLH Bridge~\cite{ref25}, HRSC2016~\cite{ref21}, LEVIR~\cite{ref24}, NWPU-VHR-10~\cite{Cheng2014_NWPUVHR10}, SIMD~\cite{Haroon2020_SIMD}, SODA-A~\cite{ref20}, ShipRSImageNet~\cite{ref22}, WHU Buildings~\cite{ref23}, and xView~\cite{ref18}. The converted files contain 175,644 training records / 3,477,064 boxes and 56,695 validation records / 1,297,548 boxes; the reported Stage-2 model uses only these original annotations, without supplemental or relabel-assisted boxes. Track A evaluates DIOR and DOTA-v2.0. Track B uses the six-source common direct/mixed ancestor protocol (36 relations, 782,132 GT) plus the separate FAIR1M grandparent diagnostic (nine relations, 290,906 GT). HRRSD~\cite{Zhang2019_HRRSD}, RSOD~\cite{Long2017_RSOD}, UCAS-AOD~\cite{Zhu2015_UCASAOD}, and VEDAI~\cite{Razakarivony2016_VEDAI} are excluded from training and used for Track-C prompt generalization; detailed splits and source roles are provided in the technical appendix.
	
	\textbf{Base detector and baselines.}
	We use WeDetect-tiny~\cite{ref40} as the base detector. Its image backbone is ConvNeXt-tiny~\cite{ref31}, its head follows a YOLO-World-style BN contrastive head, and the prompt dimension is 768. We compare against two WeDetect baselines: an official LVIS-weight zero-shot transfer baseline and a per-source flat fine-tuning baseline initialized from the official WeDetect-tiny weights. We also compare with remote sensing open-prompt or open-set baselines OpenRSD~\cite{ref1} and LAE-DINO~\cite{ref2}.

	\textbf{External-result provenance.} Track-A values for YOLO-World-L, Grounding DINO-T, and OpenRSD (text prompt) are taken from the OpenRSD paper's HBB AP50 table for DIOR-R and DOTA-v2.0, whereas those for LAE-DINO are taken from its original paper. OpenRSD does not report DOTA-v2.0 mAP for the first three rows; LAE-DINO reports DOTA-v2.0 mAP but not AP50, and unavailable entries are shown as dashes. For Track B, YOLO-World-L and Grounding DINO-T are flat-fine-tuned separately on each source's original categories for 10 epochs; OpenRSD and LAE-DINO are re-evaluated from released weights. All use method-native category prompts under shared images, labels, COCO AP implementation, and detection budget; we do not substitute \ours{} prompts. The OpenRSD Track-C score is literature-reported. LAE-DINO Track-C results are measured in this work using the official LAE-1M pretrained checkpoint and native dataset category prompts, without target-set fine-tuning. The official LAE-DINO data catalog lists RSOD under LAE-FOD/LAE-1M pretraining but does not list HRRSD, UCAS-AOD, or VEDAI; we therefore exclude RSOD only when calculating the three-source held-out comparison stated in the text. All \ours{} results and Panel~A baselines are measured in this work.

	\textbf{Efficiency protocol.} FPS is measured on an RTX 3090 with batch size 1 in FP32. The two WeDetect-tiny baselines and \ours{} text use the same cached RS153 text embeddings at inference; YOLO-World-L is likewise reparameterized with cached text. Parameter counts exclude prompt caches (the WeDetect-tiny/\ours{} text cache is 2.5~MB); full precision, latency, memory, and cache accounting is provided in the technical appendix.
	
	\textbf{Training and metrics.}
	Stage 1 trains the text path at 832 resolution, with its epoch-25 checkpoint initializing Stage 2. Stage 2 continues at 896 resolution for 15 epochs. ConvVPE is subsequently trained for 8 epochs as a frozen-body visual-prompt branch initialized from the Stage-2 hierarchy-aware text checkpoint. Track A reports COCO-style mAP, AP50, and AP75; Tracks B and C report AP50 for hierarchy queries and cross-dataset prompt transfer, respectively. Track B additionally reports conditional ancestor recall. For each ground-truth hierarchy relation $r=(i,y,a)$, let $A_r$ indicate that instance $i$ is recalled under its atomic query $y$, and let $H_r$ indicate that the same instance is recalled under ancestor query $a$. Then
	\begin{equation}
		\mathrm{CAR}_{50}=\frac{\sum_r A_rH_r}{\sum_r A_r},
		\quad \mathrm{Violation}_{50}=1-\mathrm{CAR}_{50}.
	\end{equation}
	CAR50 uses class-aware, score-ordered one-to-one matching at IoU $0.50$, method-native NMS, score $\geq0.05$, and at most 600 detections per image and query branch. We report each source, a relation-level micro aggregation, and an equal-weight source macro over DOTA-v2.0, FAIR1M, and HRSC2016; the latter two summaries expose the influence of FAIR1M's much larger relation count. Because CAR conditions on atomic success, it measures cross-granularity preservation rather than absolute coverage. Moreover, broad prompts can obtain high CAR at this permissive operating point despite false positives, so CAR is always interpreted jointly with ancestor AP50.
	
	\subsection{Track A: Atomic Category Detection}
	
	Track A evaluates atomic detection over each evaluated source's mapped subset of the 153-class vocabulary. Table~\ref{tab:main_eval} reports DIOR and DOTA-v2.0, the two shared anchors used by most external open-prompt baselines; except for the DOTA mAP column, the dataset columns report AP50.
		
		\ours{} obtains 79.7 AP50 on DIOR and 48.2 mAP / 72.3 AP50 on DOTA-v2.0. Complete eight-source text-path results are provided in the technical appendix.

	\textbf{Efficiency.} Under like-for-like FP32 inference, \ours{} is faster and lighter than OpenRSD, Grounding DINO-T, and LAE-DINO while remaining close to cached YOLO-World-L (Table~\ref{tab:main_eval}). ConvVPE requires no external visual encoder.

	\begin{figure*}[t]
		\centering
		\includegraphics[width=0.86\textwidth]{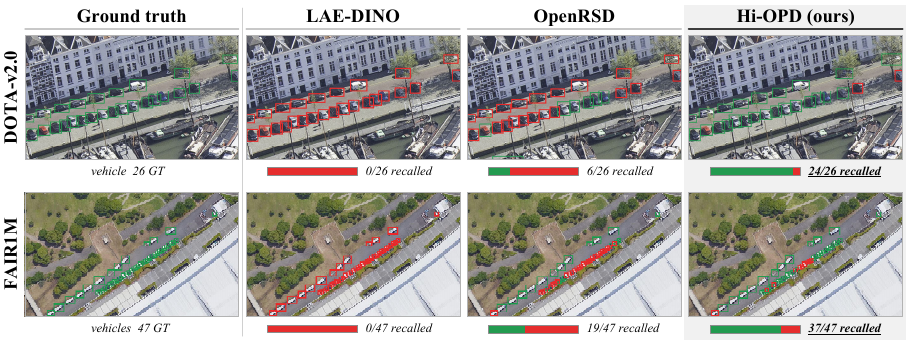}
		\caption{Track-B parent-query qualitative comparison using parent prompts only, without child-prompt expansion.}
		\label{fig:trackb_qual}
	\end{figure*}
		
	\subsection{Track B: Hierarchy-Query Recall}
	
		Track B tests whether a detector can directly use coarse prompts to recall descendant objects. It differs from merging child APs because inference neither expands child prompts nor rolls predictions up to parent labels. Table~\ref{tab:main_eval} summarizes absolute ancestor AP, and Table~\ref{tab:trackb_car} reports conditional cross-granularity preservation; the complete source-wise AP breakdown is provided in the technical appendix. The DOTA-v2.0/FAIR1M/HRSC2016 common macro is the primary cross-model subset; we additionally report the six-source AP mean for every re-evaluated public checkpoint.
	\begin{center}
		\centering
		\footnotesize
		\resizebox{\linewidth}{!}{%
			\begin{tabular}{@{}lccccc@{}}
				\toprule
				Model & DOTA & FAIR & HRSC & Common micro / src. & FAIR GP \\
				\midrule
				WeDetect-tiny (zero-shot) & 23.6 & 39.6 & 95.5 & 40.1 / 52.9 & 26.7 \\
				WeDetect-tiny (per-source ft) & 12.4 & 77.9 & 94.8 & 62.4 / 61.7 & 2.0 \\
				\textbf{\ours{} text} & \textbf{99.5} & \textbf{99.8} & 99.5 & \textbf{99.7 / 99.6} & \textbf{99.9} \\
				\ours{} visual & 97.3 & 98.1 & \textbf{99.5} & 97.9 / 98.3 & 99.0 \\
				YOLO-World-L & 9.3 & 54.8 & 4.0 & 41.4 / 22.7 & 1.6 \\
				Grounding DINO-T & 93.9 & 87.3 & 94.4 & 88.8 / 91.9 & 97.4 \\
				OpenRSD & 94.2 & 99.2 & 99.4 & 98.0 / 97.6 & 99.6 \\
				LAE-DINO & 15.7 & 94.4 & 98.4 & 75.3 / 69.5 & 7.1 \\
				\bottomrule
			\end{tabular}}
			\captionof{table}{Track-B CAR50 (\%, higher is better). Common reports relation-level micro / equal-weight source macro over DOTA-v2.0, FAIR1M, and HRSC2016; FAIR GP uses depth-2 \textit{ship}/\textit{vehicle} relations. Violation is $100-\mathrm{CAR50}$. Ancestor AP50 is reported in Table~\ref{tab:main_eval}.}
			\label{tab:trackb_car}
		\end{center}
	
		On the common three-source subset and across all six parent-query sources, \ours{} obtains 81.1 and 76.0 ancestor AP50, compared with 57.0/50.5 for OpenRSD and 33.6/27.5 for LAE-DINO. Its parent/grandparent CAR50 reaches 99.7\%/99.9\%, versus 62.4\%/2.0\% for the controlled flat WeDetect baseline. High CAR is not unique: OpenRSD reaches 98.0\%/99.6\% CAR50 but only 57.0/60.8 ancestor AP50. Thus, CAR measures conditional cross-granularity preservation, whereas AP50 measures detection accuracy; both are required. Figure~\ref{fig:trackb_qual} provides qualitative examples.
	
	\subsection{Track C: Cross-Dataset Prompt Generalization}
	
	Track C measures dataset-level open-prompt transfer rather than base/novel-category OvOD: target categories may overlap the training vocabulary, but target images are excluded from detector optimization. Text prompts are evaluated zero-shot; visual prompts use K-shot boxes from the target training split, with no target-set parameter updates or query-GT-based support selection.
	
	\begin{center}
		\centering
		\footnotesize
		\resizebox{\linewidth}{!}{%
		\begin{tabular}{@{}lcccc@{}}
			\toprule
				Dataset & \ours{} text & \ours{} visual & LAE-DINO & OpenRSD \\
				\midrule
					HRRSD & 44.2 & 42.9 & 43.9$^\ast$ & \\
					RSOD & 63.4 & 64.7 & & \\
					UCAS-AOD & 69.3 & 72.8 & 83.3$^\ast$ & \\
					VEDAI & 75.9 & 75.1 & 62.0$^\ast$ & 69.7 \\
					\midrule
					Macro & 63.1 & 63.6 & 63.1$^\ast$ & \\
			\bottomrule
		\end{tabular}}
				\captionof{table}{Track-C AP50. $^\ast$The LAE-DINO target is absent from the official LAE-1M source catalog; RSOD is omitted because it appears in LAE-FOD pretraining. Macro uses HRRSD/UCAS-AOD/VEDAI. The OpenRSD VEDAI value is literature-reported; blanks are unavailable results.}
		\label{tab:trackc}
	\end{center}
	
	On the three targets absent from LAE-1M, \ours{} text, \ours{} visual, and LAE-DINO obtain 63.1, 63.6, and 63.1 macro AP50, respectively. LAE-DINO leads on UCAS-AOD, the methods are close on HRRSD, and both \ours{} paths lead on VEDAI. On VEDAI, \ours{} text exceeds the published OpenRSD result by 6.2 points (75.9 vs.\ 69.7 AP50).
	
	\subsection{Ablation Studies}
	
	\textbf{Hierarchy-safe components.}
	Table~\ref{tab:ablation_hier} reports a cumulative four-group ablation on the original-annotation \bench{} data using epoch-15 checkpoints, 896 input, and maxDets 600; hierarchy queries use no child-prompt expansion or output roll-up. The depth-2 FAIR1M diagnostic is retained because the effect of one-way upward consistency is concentrated at the grandparent level.
	
	\begin{center}
		\centering
			\footnotesize
			\resizebox{\linewidth}{!}{%
			\begin{tabular}{@{}lccc@{}}
				\toprule
				Setting & DOTA atomic & DOTA parent & FAIR1M grandparent \\
				& mAP / AP50 & mAP / AP50 & mAP / AP50 \\
				\midrule
				C0 Flat & 47.6 / 71.4 & 3.8 / 7.2 & 15.1 / 31.6 \\
				C1 + Safe negatives & 47.6 / 71.6 & 17.3 / 35.7 & 16.8 / 34.9 \\
				C2 + PathMP & 47.6 / 71.4 & \textbf{33.1} / 71.3 & 38.6 / 66.6 \\
				C3 + Upward consistency & \textbf{48.2 / 72.3} & \textbf{33.1 / 71.5} & \textbf{45.1 / 71.4} \\
				\bottomrule
		\end{tabular}}
		\captionof{table}{Cumulative hierarchy-component ablation.}
		\label{tab:ablation_hier}
	\end{center}
	Safe negatives improve parent AP50 by 28.5 points but grandparent AP50 by only 3.3. Adding PathMP supplies the largest gains: +35.6 parent AP50 and +31.7 grandparent AP50. Upward consistency is a terminal correction for direct parents (+0.2 AP50), but remains important at depth 2, adding 4.8 grandparent AP50 while also increasing atomic AP50 by 0.9.
	
	\FloatBarrier
	
	\section{Conclusion}
	
	We presented \ours, a hierarchy-aware open-prompt detector for remote sensing images, together with \bench{} for atomic detection, hierarchy-query recall, and cross-dataset prompt generalization. Hierarchy-safe negatives, path multi-positive supervision, and upward consistency address multi-source granularity conflicts, while ConvVPE provides a lightweight detector-native visual prompt path. Joint ancestor AP and CAR evaluation shows that explicit hierarchy supervision improves both absolute coarse-query retrieval and cross-granularity preservation while retaining atomic detection.
	
	\FloatBarrier
	\bibliography{references}
	
	\FloatBarrier
	\clearpage
	\onecolumn
	\appendix
\section{Technical Appendix}

This appendix records the benchmark definition and the implementation and evaluation details that are useful for reproducibility but too detailed for the main text: taxonomy and relation coverage, source inventories, bounded leakage and license audits, auxiliary supplemental annotation details, external-baseline coverage, training hyperparameters, Track-B diagnostics, visual-prompt results, and additional qualitative examples.

\subsection{Review-Time Benchmark Definition}

\bench{} retains 153 source-observed categories as its atomic vocabulary. Atomic classes must be visually decidable at the available remote-sensing resolution, and each hierarchy edge must express a strict \textit{is-a} relation. Part-of relations, scene co-occurrence, shared function, and appearance similarity alone are excluded. Alias links encode naming equivalence or protocol-level ambiguity and do not create ancestor supervision. The hierarchy is intentionally sparse: only operationally meaningful ancestor queries are retained. Table~\ref{tab:app_benchmark_stats} summarizes its complete scale.

\begin{center}
	\centering
	\footnotesize
	\begin{tabular}{@{}lr@{}}
		\toprule
		Item & Statistics \\
		\midrule
		Atomic / hierarchy-enabled atomic classes & 153 / 108 \\
		Unique parent / grandparent nodes & 11 / 2 \\
		Direct / grandparent relations & 108 / 70 \\
		Alias groups / protected links & 3 / 11 \\
		Atomic classes also serving as ancestors & 11 \\
		Training records / GT boxes & 175,644 / 3,477,064 \\
		Validation records / GT boxes & 56,695 / 1,297,548 \\
		Effective Risk coverage (global / source-specific) & 12 / 3 sources \\
		Track-B relations (common / standalone GP) & 36 / 9 \\
		Track-B GT (common / standalone GP) & 782,132 / 290,906 \\
		\bottomrule
	\end{tabular}
	\captionof{table}{Complete benchmark scale. Track-B counts use source-specific relation entries.}
	\label{tab:app_benchmark_stats}
\end{center}

Figure~\ref{fig:app_tracks} summarizes the benchmark's three open-prompt evaluation tasks and their prompt and metric conventions.

\begin{center}
	\centering
	\includegraphics[width=0.90\textwidth]{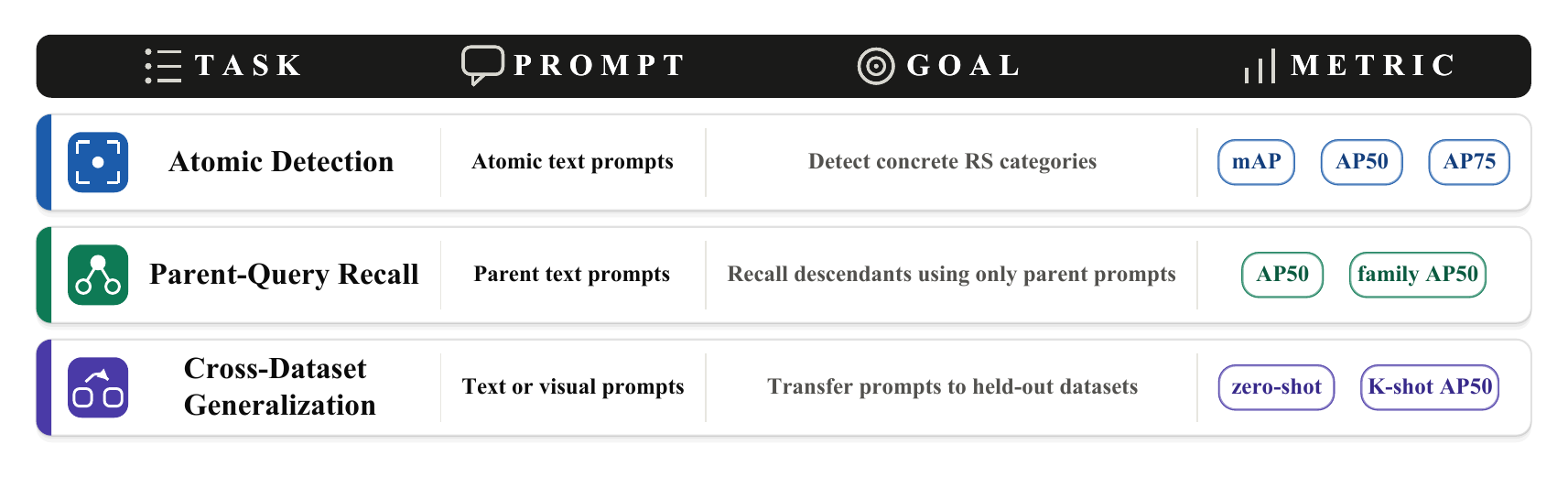}
	\captionof{figure}{Three open-prompt evaluation tasks in \bench. Track A evaluates atomic-category detection; Track B evaluates absolute and atomic-conditioned descendant recall using only hierarchy-level prompts, without child-prompt expansion or post-hoc label roll-up; and Track C evaluates text- or visual-prompt generalization on held-out datasets.}
	\label{fig:app_tracks}
\end{center}

The code-and-data appendix makes the complete definition inspectable. It contains the 153-row class table, machine-readable taxonomy, 108 direct and 70 grandparent edges, alias groups, all 246 verified source-label mappings, the effective Risk set for every source, split and tile provenance, and the exact 45 Track-B relation entries with their GT counts. Automated validation confirms unique IDs and normalized names, resolvable parent chains, no hierarchy cycles, and semantic agreement of all 45 released Track-B relations with the taxonomy. The full relation provenance check traces 1,073,038 hierarchy annotations back to their source annotation IDs, image IDs, categories, and boxes.

\begin{center}
	\centering
	\footnotesize
	\begin{tabular}{@{}lcrr@{}}
		\toprule
		Source & Protocol role & Relations & GT relations \\
		\midrule
		DOTA-v2.0 & common & 2 & 105,591 \\
		SODA-A & common & 2 & 373,905 \\
		FAIR1M & common & 9 & 290,906 \\
		HRSC2016 & common & 4 & 218 \\
		ShipRSImageNet & common & 1 & 65 \\
		SIMD & common & 18 & 11,447 \\
		\midrule
		Common subtotal &  & 36 & 782,132 \\
		FAIR1M & standalone grandparent & 9 & 290,906 \\
		\bottomrule
	\end{tabular}
	\captionof{table}{Track-B relation coverage. The FAIR1M depth-2 diagnostic is kept separate from the common direct/mixed protocol.}
	\label{tab:app_trackb_coverage}
\end{center}

\paragraph{Audit and release boundaries.}
The code-and-data appendix includes raw-image-independent metadata and validation scripts, not raw source images. All 10,838 pre-identified same-source/same-filename train/validation pairs in FAIR1M were checked at byte level; none is byte-identical. This bounded check does not establish an all-pairs perceptual near-duplicate audit or geographic isolation, which cannot be claimed for sources lacking stable geographic identity. The taxonomy has passed structural and Track-B semantic validation, but no independent second-person agreement statistic is claimed. License status is reported per source; redistribution of raw assets remains gated by the original terms, while mappings, logical split IDs, checksums, and deterministic standardized-COCO merge code are provided.

\subsection{Reproducibility Checklist Clarifications}

The novel benchmark definition is included in the code-and-data appendix: complete taxonomy, hierarchy and alias metadata, source mappings, effective Risk sets, Track-A/B/C protocol files, split provenance, bounded audits, and validation/standardized-COCO merge scripts. All source datasets are publicly available and are cited in the main paper. Upon publication, the benchmark metadata, mappings, protocols, and code will be released under a license allowing free research use. Raw source images are not redistributed; users obtain them from their official public sources and remain subject to the corresponding original terms. Thus public availability of the benchmark does not assert a new redistribution license over third-party raw imagery.

Preprocessing and experiment-code availability are also distinguished. The code-and-data appendix contains deterministic standardized-COCO merging and all benchmark validation scripts, but not every source-native download/conversion adapter or the complete training/evaluation repository; the latter is committed for public release upon publication. Random-seed documentation remains \emph{partial} because the final Track-C support-bank image/box manifest and its selection seed have not yet been recovered, although the support-origin and no-query-GT selection policy are documented. The duplicate audit is limited to exact-byte comparison of pre-identified same-source/same-filename train/validation pairs and is not claimed as an all-pairs perceptual or geographic audit.

\subsection{Data Sources and Auxiliary Supplemental Annotations}

	Table~\ref{tab:app_sources} lists the datasets used to construct \bench{} and clarifies each source's role in training or evaluation. Table~\ref{tab:app_suppann} records the separate supplemental annotation resource for partial-label sources; it is not a main-method contribution and is not used to train the headline \bench{} models. Following the dataset-inventory style of LAE-DINO, Table~\ref{tab:app_baseline_coverage} documents which Track-B sources have training-coverage evidence for the external baselines used in the main-paper Track-B comparison. The external numerical rows themselves use the shared Track-B evaluation protocol; this table only records differences in the reported training-source coverage.
	The counts in Table~\ref{tab:app_sources} are retained image/tile and box records in our converted JSON files after source-specific splitting, tiling, filtering, and label normalization; they are not native dataset totals. The mapped-label counts likewise describe labels retained after conversion into \bench{}, rather than the size of each source's original taxonomy. FAIR1M results use the validation split with available annotations rather than the official test server.

\begin{center}
	\centering
	\footnotesize
	\begin{tabularx}{\textwidth}{@{}lcccY@{}}
		\toprule
		Source & Train records / boxes & Val/test records / boxes & Mapped labels & Role \\
		\midrule
		DIOR & 11,725 / 68,070 & 11,738 / 124,441 & 20 & training; Track A validation \\
		DOTA-v2.0 & 22,021 / 534,788 & 6,612 / 159,539 & 18 & training; Track A/B validation \\
		FAIR1M & 40,701 / 795,351 & 18,528 / 369,556 & 37 & training; Track B validation only \\
		SODA-A & 36,128 / 1,126,316 & 16,576 / 583,636 & 9 & training; Track B validation only \\
		HRSC2016 & 617 / 1,748 & 438 / 1,228 & 22/23$^{\dagger}$ & training; Track B ship validation only \\
		ShipRSImageNet & 2,748 / 13,963 & 687 / 3,610 & 50 & training; Track B fine-grained ship validation only \\
		SIMD & 3,999 / 34,923 & 1,000 / 9,595 & 14 & training; Track B airport validation only \\
		WHU Buildings & 3,094 / 159,178 & 1,116 / 45,943 & 9 & main training; supplemental variant analyzed \\
		GLH Bridge & 41,601 / 83,473 & train-only & 6 & main training; supplemental variant analyzed \\
		LEVIR & 3,791 / 11,028 & train-only & 6 & main training; supplemental variant analyzed \\
		NWPU-VHR-10 & 650 / 3,896 & train-only & 10 & training only \\
		xView & 8,569 / 644,330 & train-only & 60 & training only \\
		\midrule
		\textbf{\bench{} total} & \textbf{175,644 / 3,477,064} & \textbf{56,695 / 1,297,548} & \textbf{153} & 12-source train/validation manifests \\
		\midrule
		HRRSD & -- & 10,943 / 30,043 & 13 & Track C validation \\
		RSOD & -- & 976 / 7,400 & 4 & Track C validation \\
		UCAS-AOD & -- & 1,510 / 14,597 & 2 & Track C validation \\
		VEDAI & -- & 938 / 3,001 & 3 & Track C validation \\
		\bottomrule
	\end{tabularx}
	\captionof{table}{Dataset inventory and source roles in \bench. Counts report retained records after conversion, not native dataset sizes.}
	\label{tab:app_sources}
\end{center}

\noindent$^{\dagger}$ For HRSC2016, 22/23 denotes the numbers of distinct mapped labels observed in its training/validation splits, respectively. The additional validation label is \textit{Commander}, which has one HRSC2016 validation instance. It is absent only from the HRSC2016 training split: the multi-source training set contains 120 \textit{Commander} boxes from ShipRSImageNet, so this is not an unseen category for the trained detector.

\begin{center}
	\centering
	\footnotesize
	\begin{tabularx}{\textwidth}{@{}lccYY@{}}
		\toprule
		Source & Images / boxes after supplementation & Pseudo boxes & Main relabeled classes & Exclusion / retention rule \\
		\midrule
		WHU Buildings & 3,094 / 436,455 & 277,277 & vehicle 146,354; small-vehicle 125,931; large-vehicle 4,908; a few bridge/tower/facility candidates & Keep original building GT; relabel mainly vehicle family; exclude source-implausible or overly broad categories \\
		GLH Bridge & 41,601 / 441,074 & 357,601 & small-vehicle 296,347; large-vehicle 50,066; swimming-pool 6,571; shipping-container 2,834; roundabout 1,783 & Keep original bridge GT; exclude 1,170,341 building pseudo boxes to avoid dominating training \\
		LEVIR & 3,791 / 40,894 & 29,866 & small-vehicle 21,233; building 8,581; large-vehicle 52 & Keep original airplane/ship/storage-tank labels; add building and vehicle-family pseudo boxes \\
		\bottomrule
	\end{tabularx}
	\captionof{table}{Auxiliary supplemental annotation details for partial-label sources. These data-curation details are not a main-method contribution.}
	\label{tab:app_suppann}
\end{center}

\paragraph{Exploratory reliable-annotation continuation.}
We additionally explored reliability-filtered supplemental annotations generated with SAM~3 and the final text detector, then continued training from the final Stage-2 \bench{} text checkpoint trained without supplemental annotations. This exploratory continuation is not a third method stage: it uses the same hierarchy-aware objective and serves only as a data-quality check. Under the same evaluation protocol, it did not yield consistent gains on the main DIOR and DOTA-v2.0 anchors. We therefore report all headline text-path results from that Stage-2 epoch-15 checkpoint and treat the supplemental annotations only as an auxiliary analysis rather than a method contribution.

\paragraph{Two roles of $\mathrm{Risk}(s)$.}
$\mathrm{Risk}(s)$ is an unsafe-negative set with a primary training role and an auxiliary data-curation role. The policy used by the original-annotation models applies one global \textit{building} protection to all twelve sources and adds source-specific sets for GLH Bridge, WHU Buildings, and LEVIR. During headline hierarchy-aware training, effective Risk members are removed from the negative prompt pool for source $s$ because they may occur in its images but are not exhaustively annotated. This exclusion does not assign positive labels to unannotated objects; it only prevents those categories from being optimized as negatives. The code-and-data appendix includes the complete effective set for all twelve sources, its scope, and its action. In the auxiliary supplemental-annotation analysis, we use the responses preserved for $\mathrm{Risk}(s)$ categories to mine potential missing boxes. Table~\ref{tab:app_risk_mining} compares candidates mined on WHU with and without source-risk exclusion: vehicle and small-vehicle candidates increase substantially at low detector-score thresholds when the exclusion is enabled. Because WHU has no vehicle ground truth, these counts are a candidate-recall proxy rather than Recall@IoU or final AP. The resulting supplemental annotations are not used by the headline models, and candidate-count growth is not claimed as final-AP improvement; the exploratory continuation above did not yield consistent gains on the main anchors.

\begin{center}
	\centering
	\footnotesize
	\begin{tabular}{@{}clrrr@{}}
		\toprule
		Score threshold & Category & No-risk & With-risk & $\Delta$ \\
		\midrule
		0.005 & vehicle & 276 & 1,489 & +1,213 \\
		0.005 & small-vehicle & 105 & 738 & +633 \\
		0.01 & vehicle & 48 & 437 & +389 \\
		0.01 & small-vehicle & 31 & 230 & +199 \\
		0.03 & vehicle & 3 & 54 & +51 \\
		0.05 & vehicle & 1 & 25 & +24 \\
		\bottomrule
	\end{tabular}
	\captionof{table}{Comparison of supplementary-annotation candidates mined on WHU with and without $\mathrm{Risk}(s)$. Scores are detector ranking scores, not calibrated probabilities.}
	\label{tab:app_risk_mining}
\end{center}

\begin{center}
	\centering
	\footnotesize
	\begin{tabularx}{\textwidth}{@{}lcccY@{}}
		\toprule
		Track-B source & \ours{} & OpenRSD & LAE-DINO & Evidence summary \\
		\midrule
		DOTA-v2.0 & seen & seen & seen & OpenRSD reports DOTA-v2.0 in its source inventory; LAE-DINO reports DOTAv2.0 benchmark/fine-tuning. \\
		SODA-A & seen & not evidenced & not evidenced & OpenRSD reports SODA-A only in cross-dataset evaluation; LAE-DINO does not list SODA/SODA-A in its source inventory. \\
		FAIR1M & seen & seen & seen & OpenRSD includes FAIR1M-2.0; LAE-DINO lists FAIR1M in its LAE-1M source tables. \\
		HRSC2016 & seen & seen & seen & Both OpenRSD and LAE-DINO include HRSC2016 in their source inventories. \\
		ShipRSImageNet & seen & seen & not evidenced & OpenRSD includes ShipRSImageNet; LAE-DINO does not list ShipRSImageNet in its source inventory. \\
		SIMD & seen & not evidenced & not evidenced & Neither OpenRSD nor LAE-DINO lists SIMD as a training source. \\
		\bottomrule
	\end{tabularx}
	\captionof{table}{Training-source coverage evidence for Track-B external diagnostics, based on the source inventories and evaluation descriptions in OpenRSD and LAE-DINO; full citations are provided in the main paper. This table concerns training-source coverage only: all numerical comparisons in the main paper use the shared Track-B evaluation images, labels, splits, COCO AP implementation, and detection budget, while each method retains its native published prompts. ``Not evidenced'' means that we did not find training-source evidence in the corresponding paper; those sources are excluded from the three-method common macro.}
	\label{tab:app_baseline_coverage}
\end{center}

\subsection{Training Hyperparameters}

Table~\ref{tab:app_hyper} summarizes the training recipe. Following the main-text convention, Stage 1 pretrains the multi-source text detector, Stage 2 continues it with hierarchy-aware supervision, and the ConvVPE visual branch is trained separately with a frozen detector body.

\begin{center}
	\centering
	\footnotesize
	\begin{tabularx}{\textwidth}{@{}lYYY@{}}
		\toprule
		Parameter & Stage 1 text pretraining & Stage 2 hierarchy-aware text & ConvVPE visual branch \\
		\midrule
		Training recipe & multi-source cached text & hierarchy-safe continuation & frozen-body visual prompt learning \\
		Initialization & public WeDetect weights & Stage 1 epoch 25 & Stage 2 epoch 15 \\
		Input size & 832 & 896 & 896 \\
		Training epochs & 36 (select epoch 25) & 15 & 8 \\
		Frozen modules & -- & -- & backbone / neck / non-BN head / text cache \\
		Learning rate & 2e-5 & 5e-6 & 2.5e-5 \\
		Weight decay & 0.025 & 0.025 & 0.05 \\
		loss bbox / dfl / cls & 7.5 / 0.375 / 0.5 & same as Stage 1 & same as Stage 1 \\
		Support samples $M$ & -- & -- & train: \{1,2,3,5\}; eval: 5 \\
		Support dropout & -- & -- & 0.10 \\
		$\lambda_{\mathrm{align}}$ & -- & -- & 1.0 \\
		Ancestor weights $\beta_1/\beta_2$ & -- & 0.8 / 0.8 & through shared head \\
		Upward margin $m$ / $\lambda_{\mathrm{up}}$ & -- & 0.0 / 0.05 & through shared head \\
		Detach child logit & -- & true & through shared head \\
		Hierarchy supervision & -- & path multi-positive, depth $\leq2$; zero-margin one-way ordering & through shared head \\
		Training prompt classes & 80 per image & 80 per image & 80 per image \\
		\bottomrule
	\end{tabularx}
	\captionof{table}{Stage-wise training hyperparameters.}
	\label{tab:app_hyper}
\end{center}

\paragraph{Compute environment.}
All training and evaluation experiments were conducted on a server with two AMD EPYC 7302 CPUs (32 physical cores and 64 threads), 256 GB of system memory, and eight NVIDIA GeForce RTX 3090 GPUs with 24 GB memory each, running Ubuntu 24.04.1 LTS. Hi-OPD and the controlled MMDetection-based baselines used Python 3.10.20, PyTorch 2.5.1+cu124 with CUDA 12.4 and cuDNN 9.1, TorchVision 0.20.1, MMEngine 0.10.7, MMCV 2.1.0, MMDetection 3.3.0, OpenCV 4.13.0, Transformers 4.57.1, and timm 1.0.27. YOLO-World-L used its released environment with Python 3.9.25, PyTorch 1.13.1+cu117, MMEngine 0.10.3, MMCV 2.0.0, MMDetection 3.0.0, MMYOLO 0.6.0, and YOLO-World 0.1.0. Distributed training used eight GPUs, whereas all efficiency measurements used one RTX 3090 with batch size 1.

Stage-1 validation uses the eight source-specific splits with available in-domain validation annotations---DIOR, DOTA-v2.0, SODA-A, WHU Buildings, FAIR1M, HRSC2016, ShipRSImageNet, and SIMD---comprising 56,695 records and 1,297,548 boxes in total. Validation is scheduled every five epochs. The configured automatic \texttt{save\_best} rule monitors the equal-weight mean of DIOR and DOTA-v2.0 Track-A AP50, the two shared main-table anchors, and selects epoch 25: it reaches 72.6, compared with 67.1, 71.6, 71.6, 71.9, 71.5, and 71.9 at epochs 10, 15, 20, 30, 35, and the final epoch 36, respectively. The corresponding equal-weight anchor mAP also peaks at epoch 25 (50.0).

\subsection{Controlled Hierarchy-Component Ablation}

The four cumulative groups in Table~\ref{tab:app_hierarchy_ablation} use the same original-annotation \bench{} training and validation data, epoch-15 checkpoints, 896 input, maxDets 600, \texttt{nms\_pre=30000}, and cached-text evaluation implementation. C0 uses flat training; C1 adds hierarchy-safe negatives; C2 additionally adds path multi-positive supervision (PathMP); and C3 adds detached one-way upward consistency.

\begin{center}
	\centering
	\footnotesize
	\begin{tabular}{@{}lccc@{}}
		\toprule
		Setting & DOTA atomic & DOTA parent & FAIR1M grandparent \\
		& mAP / AP50 & mAP / AP50 & mAP / AP50 \\
		\midrule
		C0 Flat & 47.6 / 71.4 & 3.8 / 7.2 & 15.1 / 31.6 \\
		C1 + Safe negatives & 47.6 / 71.6 & 17.3 / 35.7 & 16.8 / 34.9 \\
		C2 + PathMP & 47.6 / 71.4 & \textbf{33.1} / 71.3 & 38.6 / 66.6 \\
		C3 + Upward consistency & \textbf{48.2 / 72.3} & \textbf{33.1 / 71.5} & \textbf{45.1 / 71.4} \\
		\bottomrule
	\end{tabular}
	\captionof{table}{Original-annotation hierarchy-component ablation under the unified epoch-15 protocol.}
	\label{tab:app_hierarchy_ablation}
\end{center}

Safe negatives contribute +28.5 parent AP50 but only +3.3 grandparent AP50. PathMP is the main source of hierarchy-query improvement, adding +35.6 parent and +31.7 grandparent AP50 over C1. The final upward-consistency term changes direct-parent AP50 by only +0.2 but adds +4.8 grandparent AP50 and +0.9 atomic AP50. This supports its role as a zero-margin terminal ordering correction whose clearest effect appears on depth-2 retrieval.

\subsection{Unsafe-Negative Exposure and Additional Track-B Results}

We report the \emph{unsafe-negative exposure rate} of flat sampling: the fraction of sampled flat negatives that fall into the policy-defined ancestor, descendant, alias, or source-risk sets. This is not a GT-verified false-negative rate. Table~\ref{tab:app_false} reports the auxiliary diagnostic on ten of the twelve training sources (166,425 images); xView and NWPU-VHR-10 remain part of \bench{} training but are not included. Within each source, the listed value is constant by construction because the protected vocabulary and prompt budget are source-level, which explains why the previous mean/median/max summaries were identical. The diagnostic-subset value is image-weighted across sources.
Table~\ref{tab:app_trackb_diag} further breaks down text-path Track-B performance by direct parent source, and Table~\ref{tab:app_trackb_crossmodel} gives the corresponding cross-model AP50 comparison. Table~\ref{tab:app_trackb_gp} reports the FAIR1M grandparent diagnostic, and Table~\ref{tab:app_trackc_text} gives the text zero-shot Track-C source-wise results. Tables~\ref{tab:app_visual_tracka}, \ref{tab:app_visual_trackb}, \ref{tab:app_visual_trackb_gp}, and \ref{tab:app_visual_trackc} report the complete final ConvVPE evaluation on Track A, direct-parent Track B, the standalone Track-B grandparent diagnostic, and Track C, respectively.

\noindent\begin{minipage}[t]{0.485\textwidth}
	\vspace{0pt}
	\centering
	\footnotesize
		\begin{tabularx}{\linewidth}{@{}Yrr@{}}
			\toprule
			Source & Images & Exposure \\
			\midrule
			Single-class sources & 48,486 & 1.00 \\
			DIOR & 11,725 & 0.81 \\
			DOTA-v2.0 & 22,021 & 0.73 \\
			FAIR1M & 40,701 & 0.72 \\
			SODA-A & 36,128 & 0.67 \\
			HRSC2016 & 617 & 0.26 \\
			SIMD & 3,999 & 0.10 \\
			ShipRSImageNet & 2,748 & 0.08 \\
			\midrule
			\textbf{Diagnostic subset} & \textbf{166,425} & \textbf{0.74} \\
			\bottomrule
		\end{tabularx}
		\captionof{table}{Policy-derived unsafe-negative exposure under flat sampling on the ten-source diagnostic subset.}
	\label{tab:app_false}
\end{minipage}\hfill
\begin{minipage}[t]{0.485\textwidth}
	\vspace{0pt}
	\centering
	\footnotesize
		\begin{tabularx}{\linewidth}{@{}YYrr@{}}
			\toprule
			Source & Parent prompt & GT & AP50 \\
			\midrule
			DOTA-v2.0 & vehicle & 105,591 & 71.5 \\
			SODA-A & vehicle & 373,905 & 79.2 \\
			FAIR1M & parent classes & 290,906 & 74.6 \\
			HRSC2016 & parent classes & 218 & 97.2 \\
			ShipRSImageNet & Passenger Ship & 65 & 42.5 \\
				SIMD & parent classes (depth-1) & 11,447 & 91.2 \\
			\midrule
			Macro & all sources & -- & 76.0 \\
			\bottomrule
		\end{tabularx}
		\captionof{table}{Track-B direct-parent diagnostics for the epoch-15 Stage-2 \bench{} text model trained without supplemental annotations. The standalone FAIR1M grandparent diagnostic is reported in Table~\ref{tab:app_trackb_gp}.}
	\label{tab:app_trackb_diag}
\end{minipage}

\medskip
\begin{center}
	\centering
	\footnotesize
	\begin{tabular}{@{}llccc@{}}
		\toprule
		Source & Ancestor prompt & \ours{} text & OpenRSD & LAE-DINO \\
		\midrule
		DOTA-v2.0 & vehicle & 71.5 & 61.0 & 11.6 \\
		SODA-A & vehicle & 79.2 & 67.4$^\ast$ & 1.6$^\ast$ \\
		FAIR1M & cargo/vehicle & 74.6 & 64.0 & 59.2 \\
		HRSC2016 & ship family & 97.2 & 46.1 & 30.1 \\
		ShipRSImageNet & Passenger Ship & 42.5 & 14.2$^\ast$ & 1.2$^\ast$ \\
		SIMD & direct parents & 91.2 & 50.4$^\ast$ & 61.4$^\ast$ \\
		\midrule
		Common macro & DOTA/FAIR1M/HRSC2016 & \textbf{81.1} & 57.0 & 33.6 \\
		All-source macro & six sources & \textbf{76.0} & 50.5 & 27.5 \\
		\midrule
		FAIR1M grandparent & ship/vehicle & \textbf{71.4} & 60.8 & 20.9 \\
		\bottomrule
	\end{tabular}
	\captionof{table}{Complete Track-B hierarchy-query AP50 comparison. $^\ast$The corresponding released OpenRSD or LAE-DINO model has no evidenced training coverage for that source and is evaluated directly from its public checkpoint. The common macro uses the three sources with evidenced coverage for all three methods.}
	\label{tab:app_trackb_crossmodel}
\end{center}

For CAR50, DOTA-v2.0, FAIR1M, and HRSC2016 contain 105,591, 290,906, and 218 direct-parent ground-truth relations, respectively; the FAIR1M grandparent diagnostic contains 290,906 depth-2 relations. The CAR denominator is the model-dependent subset of these relations recalled under atomic queries. Consequently, the relation-micro summary is strongly influenced by FAIR1M, while the equal-source macro in the main paper weights the three sources equally. Both summaries and all three source-wise CAR50 values are therefore reported.

\medskip
\noindent\begin{minipage}[t]{0.485\textwidth}
	\vspace{0pt}
	\centering
	\footnotesize
	\begin{tabular}{@{}lrrr@{}}
		\toprule
		Grandparent prompt & GT & mAP & AP50 \\
		\midrule
		ship & 16,820 & 38.1 & 50.4 \\
		vehicle & 274,086 & 52.0 & 92.3 \\
		\midrule
		Overall & 290,906 & 45.1 & 71.4 \\
		\bottomrule
	\end{tabular}
	\captionof{table}{FAIR1M grandparent diagnostic for text prompts under maxDets 600. Overall is the equal-weight macro over the \textit{ship} and \textit{vehicle} prompts; it is not part of the six-source direct-parent macro.}
	\label{tab:app_trackb_gp}
\end{minipage}\hfill
\begin{minipage}[t]{0.485\textwidth}
	\vspace{0pt}
	\centering
	\footnotesize
	\begin{tabular}{@{}lrrrr@{}}
		\toprule
		Source & Classes & mAP & AP50 & AP75 \\
		\midrule
		HRRSD & 13 & 30.5 & 44.2 & 35.5 \\
		RSOD & 4 & 35.8 & 63.4 & 34.1 \\
		UCAS-AOD & 2 & 29.7 & 69.3 & 21.4 \\
		VEDAI & 3 & 54.6 & 75.9 & 64.3 \\
		\midrule
		4-source macro &  & 37.7 & 63.2 & 38.8 \\
		\bottomrule
	\end{tabular}
	\captionof{table}{Track-C text zero-shot source-wise results for the epoch-15 Stage-2 \bench{} model trained without supplemental annotations.}
	\label{tab:app_trackc_text}
\end{minipage}

\medskip
The LAE-1M coverage designation follows the official LAE-DINO data catalog (\url{https://github.com/jaychempan/LAE-DINO#dataset}), which lists RSOD under LAE-FOD but does not list HRRSD, UCAS-AOD, or VEDAI.
\begin{center}
	\centering
	\footnotesize
	\begin{tabular}{@{}lccc@{}}
		\toprule
		Source & LAE-DINO mAP & AP50 & AP75 \\
		\midrule
		HRRSD$^\ast$ & 29.5 & 43.9 & 33.8 \\
		RSOD & & & \\
		UCAS-AOD$^\ast$ & 40.0 & 83.3 & 37.0 \\
		VEDAI$^\ast$ & 38.7 & 62.0 & 42.7 \\
		\midrule
		Macro$^\ast$ & 36.1 & 63.1 & 37.8 \\
		\bottomrule
	\end{tabular}
	\captionof{table}{LAE-DINO Track-C evaluation using the official LAE-1M pretrained checkpoint, native category prompts, and no target-set fine-tuning. $^\ast$The target is absent from the official LAE-1M source catalog. The RSOD row is intentionally left blank because RSOD appears in LAE-FOD pretraining and is excluded from comparison. Macro is computed over HRRSD, UCAS-AOD, and VEDAI.}
	\label{tab:app_trackc_laedino}
\end{center}

\subsection{Complete Text-Path Track-A Results}

Table~\ref{tab:app_text_tracka_full} reports the final Stage-2 epoch-15 text checkpoint under the full Track-A protocol (896 input, COCO bbox evaluation, and maxDets 600). Each source is evaluated on its mapped subset of the 153-class vocabulary. The DIOR and DOTA-v2.0 AP50 values match the main-paper headline results; validation-set sizes are listed in Table~\ref{tab:app_sources}. A dash indicates that the source contains no valid COCO ground truth at that object scale.

\begin{center}
	\centering
	\footnotesize
	\begin{tabular}{@{}lrrrrrrr@{}}
		\toprule
		Source & Cats. & mAP & AP50 & AP75 & APs & APm & APl \\
		\midrule
		DIOR & 20 & 57.2 & 79.7 & 62.5 & 19.2 & 45.3 & 77.0 \\
		DOTA-v2.0 & 18 & 48.2 & 72.3 & 54.2 & 27.0 & 50.3 & 59.4 \\
		SODA-A & 9 & 39.7 & 81.7 & 44.8 & 39.6 & 52.8 & -- \\
		WHU Buildings & 1 & 78.3 & 96.1 & 90.5 & 52.8 & 84.8 & 91.5 \\
		FAIR1M & 37 & 28.8 & 41.3 & 32.1 & 12.3 & 25.9 & 39.0 \\
		HRSC2016 & 23 & 70.0 & 77.2 & 76.3 & -- & 51.3 & 70.8 \\
		ShipRSImageNet & 50 & 49.9 & 63.6 & 55.5 & 15.8 & 32.3 & 54.2 \\
		SIMD & 14 & 69.1 & 85.3 & 80.2 & 26.6 & 56.0 & 74.8 \\
		\midrule
		\textbf{8-source macro} & -- & \textbf{55.2} & \textbf{74.7} & \textbf{62.0} & \textbf{27.6}$^\ast$ & \textbf{49.8} & \textbf{66.7}$^\ast$ \\
		\bottomrule
	\end{tabular}
	\captionof{table}{Complete Track-A results for the final text path. Source macros weight the eight sources equally. $^\ast$APs excludes HRSC2016 and APl excludes SODA-A because the corresponding COCO scale metric is undefined.}
	\label{tab:app_text_tracka_full}
\end{center}

\subsection{Final ConvVPE Results}

All visual results below use the final ConvVPE checkpoint trained for eight epochs with a frozen detector body, initialized from the final Stage-2 epoch-15 text checkpoint, trained only on the original converted \bench{} annotations without supplemental or relabel-assisted boxes, and evaluated with $M=5$. It obtains 75.2 AP50 on DIOR and 46.1 mAP / 69.5 AP50 on DOTA-v2.0, giving a 72.4 two-source AP50 average. Earlier internal component and support-number studies came from a relabel-assisted branch initialized from its epoch-10 text checkpoint and also used a different ConvVPE checkpoint and support bank. The DIOR/DOTA validation annotations and evaluation protocol were byte-identical, so the discrepancy was not caused by evaluation data; nevertheless, those results do not constitute ablations of the final original-annotation model and are excluded from this paper. Because the component and $M$-sweep analyses have not yet been repeated with the final original-annotation checkpoint, we omit the older numbers rather than mix training branches. Every numerical result measured in this work is a single evaluation of a fixed checkpoint; no repeated-seed mean or variance is claimed.

\noindent\begin{minipage}[t]{0.485\textwidth}
	\vspace{0pt}
	\centering
	\footnotesize
	\begin{tabular}{@{}lrr@{}}
		\toprule
		Source & mAP & AP50 \\
		\midrule
		DIOR & 53.9 & 75.2 \\
		DOTA-v2.0 & 46.1 & 69.5 \\
		SODA-A & 35.7 & 75.9 \\
		WHU Buildings & 78.0 & 96.0 \\
		FAIR1M & 25.3 & 35.5 \\
		HRSC2016 & 54.8 & 60.3 \\
		ShipRSImageNet & 30.2 & 38.5 \\
		SIMD & 57.7 & 70.9 \\
		\midrule
		\textbf{8-source macro} & \textbf{47.7} & \textbf{65.2} \\
		\bottomrule
	\end{tabular}
	\captionof{table}{Complete Track-A visual K-shot results for the final ConvVPE branch (epoch 8, 896 input, $M=5$).}
	\label{tab:app_visual_tracka}
\end{minipage}\hfill
\begin{minipage}[t]{0.485\textwidth}
	\vspace{0pt}
	\centering
	\footnotesize
	\begin{tabular}{@{}lrr@{}}
		\toprule
		Source / query & mAP & AP50 \\
		\midrule
		DOTA-v2.0 / vehicle & 31.3 & 69.8 \\
		SODA-A / vehicle & 20.9 & 78.2 \\
		FAIR1M / parent classes & 40.3 & 66.7 \\
		HRSC2016 / parent classes & 45.6 & 49.3 \\
		ShipRSImageNet / Passenger Ship & 4.5 & 6.6 \\
		SIMD / parent classes (depth-1) & 71.9 & 87.8 \\
		\midrule
		Common macro (DOTA/FAIR1M/HRSC) & \textbf{39.1} & \textbf{61.9} \\
		All-source macro & \textbf{35.8} & \textbf{59.7} \\
		\bottomrule
	\end{tabular}
	\captionof{table}{Complete Track-B direct-parent results for the final ConvVPE branch (epoch 8, 896 input, $M=5$). The common macro matches the DOTA-v2.0, FAIR1M, and HRSC2016 coverage used for external-method comparison.}
	\label{tab:app_visual_trackb}
\end{minipage}

\medskip
\noindent\begin{minipage}[t]{0.485\textwidth}
	\vspace{0pt}
	\centering
	\footnotesize
	\begin{tabular}{@{}lrr@{}}
		\toprule
		Grandparent query & mAP & AP50 \\
		\midrule
		ship & 20.0 & 28.4 \\
		vehicle & 45.0 & 86.0 \\
		\midrule
		Overall & \textbf{32.5} & \textbf{57.2} \\
		\bottomrule
	\end{tabular}
	\captionof{table}{FAIR1M grandparent visual K-shot diagnostic for the final ConvVPE branch (epoch 8, 896 input, $M=5$, maxDets 600). Overall is the equal-weight macro over the \textit{ship} and \textit{vehicle} prompts and is separate from the six-source direct-parent macro.}
	\label{tab:app_visual_trackb_gp}
\end{minipage}\hfill
\begin{minipage}[t]{0.485\textwidth}
	\vspace{0pt}
	\centering
	\footnotesize
	\begin{tabular}{@{}lrrr@{}}
		\toprule
		Source & mAP & AP50 & AP75 \\
		\midrule
		HRRSD & 29.2 & 42.9 & 33.9 \\
		RSOD & 36.9 & 64.7 & 35.8 \\
		UCAS-AOD & 31.2 & 72.8 & 22.4 \\
		VEDAI & 54.2 & 75.1 & 64.1 \\
		\midrule
		\textbf{4-source macro} & \textbf{37.9} & \textbf{63.9} & \textbf{39.1} \\
		\bottomrule
	\end{tabular}
	\captionof{table}{Complete Track-C visual K-shot results for the final ConvVPE branch (epoch 8, 896 input, $M=5$).}
	\label{tab:app_visual_trackc}
\end{minipage}

\subsection{Text and Visual Prompt Paths}

Table~\ref{tab:app_text_visual} compares the final text and visual prompt paths on the two Track-A sources reported in the main table. The visual path uses only detector-native support features.

\begin{center}
	\centering
	\footnotesize
		\begin{tabularx}{0.82\textwidth}{@{}YYcc@{}}
			\toprule
			Path & Prompt source & DIOR mAP / AP50 & DOTA mAP / AP50 \\
			\midrule
			Final text path (Stage 2) & XLM-R text embedding & 57.2 / 79.7 & 48.2 / 72.3 \\
			Final ConvVPE visual & 5-shot support bank & 53.9 / 75.2 & 46.1 / 69.5 \\
			\bottomrule
		\end{tabularx}
	\captionof{table}{Text path versus final ConvVPE visual prompt path on the main Track-A sources (epoch 8, 896 input, $M=5$).}
	\label{tab:app_text_visual}
\end{center}

\subsection{Deployment Parameter and Cache Accounting}

Table~\ref{tab:efficiency} reports the full inference speed and memory measurements behind the FPS column of the main results table. Table~\ref{tab:app_online_params} separates non-cache online evaluation parameters from offline prompt encoders, and Table~\ref{tab:app_cache} reports prompt/support cache footprint.

\begin{center}
	\centering
	\footnotesize
	\begin{tabular}{@{}llrrr@{}}
		\toprule
		Model & Mode & FPS & ms/img & Memory \\
		\midrule
		YOLO-World-L & cached FP32 & \textbf{48.539} & \textbf{20.602} & 687.7 MB \\
		\ours{} text & AMP-safe & 44.107 & 22.672 & \textbf{294.2 MB} \\
		\ours{} visual M=5 & AMP-safe & 42.299 & 23.641 & 382.0 MB \\
		\ours{} text & FP32 & 37.914 & 26.376 & 354.4 MB \\
		\ours{} visual M=5 & FP32 & 36.369 & 27.496 & 442.1 MB \\
		WeDetect-tiny & online XLM-R FP32 & 30.589 & 32.692 & 1416.9 MB \\
		OpenRSD & FP32 & 29.603 & 33.781 & 517.8 MB \\
		Grounding DINO-T & FP32 & 9.7 & 103.1 & 1030 MB \\
		LAE-DINO & FP32 & 7.9 & 126.6 & 1057 MB \\
		\bottomrule
	\end{tabular}
	\captionof{table}{Inference speed and memory on RTX 3090, batch size 1. FP32 rows provide the like-for-like comparison used in the main results table; AMP-safe rows are the \ours{} deployment mode.}
	\label{tab:efficiency}
\end{center}

\begin{center}
	\centering
	\footnotesize
	\begin{tabularx}{\textwidth}{@{}lYYY@{}}
		\toprule
		Model & Online / eval parameters & Text or language path & Visual / prompt path \\
		\midrule
		\ours{} text cached & 38.078M & 0 online; offline XLM-R 278.634M & ConvNeXt 28.589M \\
		\ours{} visual + ConvVPE & 41.510M & 0 online & ConvNeXt 28.589M + ConvVPE 3.432M \\
		WeDetect-tiny online & 316.712M & XLM-R 278.634M online & ConvNeXt 28.589M \\
		YOLO-World-L & 110.312M before reparameterization & CLIP text 63.428M & YOLO backbone 19.832M \\
		OpenRSD & 67.227M eval / 76.752M checkpoint & support text cached & DINOv2 offline only \\
		Grounding DINO-T & 173.007M & BERT 108.892M & Swin-T + fusion $\approx$64.115M \\
		LAE-DINO & 181.567M & BERT 108.892M & Swin-T + fusion $\approx$72.675M \\
		\bottomrule
	\end{tabularx}
	\captionof{table}{Non-cache parameter accounting for online evaluation. Cached prompt or support embeddings are excluded here and summarized separately in Table~\ref{tab:app_cache}.}
	\label{tab:app_online_params}
\end{center}

\begin{center}
	\centering
	\footnotesize
	\begin{tabularx}{\textwidth}{@{}lYY@{}}
		\toprule
		Cache & Content & Size \\
		\midrule
		\ours{} text cache & 831 prompts $\times$ 768 float32 & 2.5 MB \\
		\ours{} visual support cache & 153 classes, at most 200 supports per class, 768 float32 & 71 MB \\
		YOLO-World one-class cache & $1 \times 512$ & 2 KB \\
		YOLO-World RS153 cache & $153 \times 512$ & 0.30 MiB \\
		OpenRSD DOTA2 support cache & 18 classes with 16 text prompts $\times$ 768 and 50 visual prompts $\times$ 1024 & 4.0 MB \\
		OpenRSD negative support cache & 237 negative support classes & 59 MB \\
		\bottomrule
	\end{tabularx}
	\captionof{table}{Prompt and support cache footprint used by cached-inference variants.}
	\label{tab:app_cache}
\end{center}

\subsection{Additional Qualitative Examples}

Figure~\ref{fig:tracka_qual} complements the parent-query qualitative comparison in the main paper with atomic Track-A examples on DIOR and DOTA-v2.0. We keep this larger visual comparison in the appendix so that the main text can foreground the primary quantitative tables and the hierarchy-specific Track-B examples.

\begin{center}
	\centering
	\includegraphics[width=0.88\textwidth]{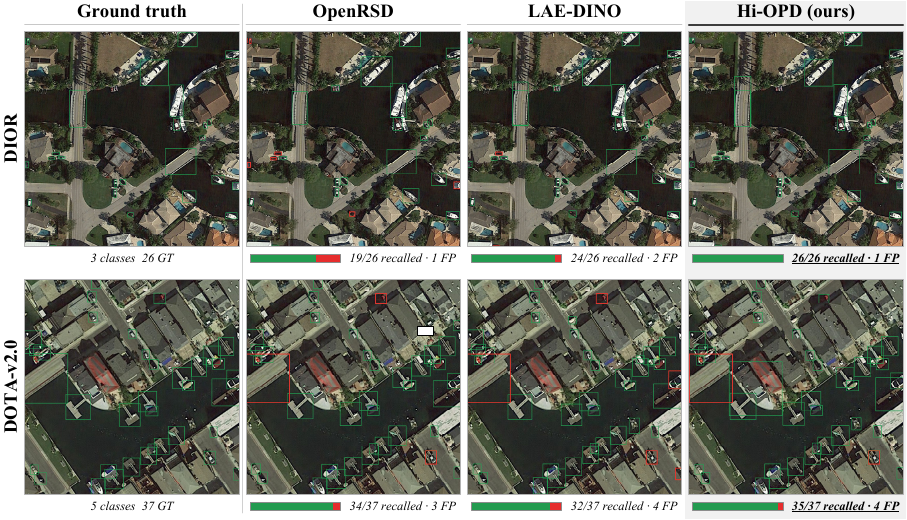}
	\captionof{figure}{Track-A atomic detection qualitative comparison. Green boxes denote detections and red boxes denote missed ground-truth objects in method panels.}
	\label{fig:tracka_qual}
\end{center}

	\end{document}